\documentclass[12pt]{article}

\usepackage{graphicx}
\usepackage{amsmath}
\usepackage{url}
\usepackage{hyperref}

\begin{document}
\title{Enhancing LLM Evaluations: The Garbling Trick}
\author{William F. Bradley\thanks{Mirabolic Consulting}}
\maketitle

\begin{abstract}
As large language models (LLMs) become increasingly powerful, traditional evaluation metrics tend to saturate, making it challenging to distinguish between models. We propose a general method to transform existing LLM evaluations into a series of progressively more difficult tasks. These enhanced evaluations emphasize reasoning capabilities and can reveal relative performance differences that are not apparent in the original assessments.

To demonstrate the effectiveness of our approach, we create a new multiple-choice test corpus, extend it into a family of evaluations, and assess a collection of LLMs. Our results offer insights into the comparative abilities of these models, particularly highlighting the differences between base LLMs and more recent ``reasoning'' models.
\end{abstract}

\section{Introduction}
Significant advancements in AI and machine learning have often progressed in tandem with the development of evaluation tools. For instance, MNIST~\cite{lecun2010mnist} spurred the creation of LeNet~\cite{lecun1998gradient}, ImageNet~\cite{deng2009imagenet} led to the development of AlexNet~\cite{krizhevsky2012imagenet}, and CASP~\cite{moult2020critical} motivated AlphaFold~\cite{jumper2021highly}. However, it is common for performance on a given test to reach saturation—for example, modern accuracy on MNIST exceeds 99.9\%~\cite{an2020ensemble}, rendering it less useful for comparing models or driving research.

Despite being a relatively young field, the evaluation of large language models (LLMs) faces similar challenges. Multiple-choice tests are a popular evaluation method (e.g., MMLU~\cite{hendrycks2020measuring}, MMLU-PRO~\cite{wang2024mmlu}, HellaSwag~\cite{zellers2019hellaswag}, HELM~\cite{liang2022holistic}, GSM8K~\cite{cobbe2021training}), but saturation is already apparent. As of October 2024, top-performing models have achieved MMLU scores of 92.3\%~\cite{openai2023learning}, MMLU-PRO scores of 91.0\%~\cite{chen2024tweet}, HellaSwag scores of 96.1\%, HELM (Lite) scores of 95.9\%~\cite{helm2024holistic}, and GSM8K scores exceeding 95\%~\cite{mirzadeh2024gsm}. OpenAI has even argued~\cite{openai2023learning} that the GSM8K metric is no longer effective for differentiating models.

Why not simply introduce new, more challenging tests? Beyond the difficulty of creating well-designed assessments, the greatest value of an evaluation framework lies in how it guides research in the field. MNIST, ImageNet, and CASP were beneficial because they took years to saturate, providing clear objectives for numerous research groups during that time. Ideally, a test should be sufficiently challenging to engage the field for several years, not just a few months.

In this paper, we present a straightforward method for transforming any text-based evaluation into a family of increasingly challenging tests. This is a meta-evaluation technique: it takes an existing evaluation framework and converts it into a new one. To demonstrate our method, we generate a new evaluation framework and then extend it into a harder family of tests. Even this simple example reveals non-obvious differences in performance among models.

\section{The Garbling Trick}
The concept we explore is straightforward and can be summarized in one sentence:
\begin{quote}
Given a text-based evaluation method, randomly garble the text and observe how varying the garbling rate impacts the results.
\end{quote}
More specifically, suppose we have an evaluation comprising a set of problems. Each problem includes some context, a question about that context, and a set of possible answers. The evaluation framework assigns a score $s$ to the responses (e.g., the percentage of correct answers). We garble each character in the context with a probability $p$ and then compute $s(p)$. While the original evaluation yields a single score, the garbled evaluation produces a curve as a function of $p$, ranging from $p=0$ to $p=1$, which we refer to as the \emph{score curve}. Finally, we focus the evaluation on a challenging subset of problems, a process detailed in Section~\ref{sec:core}.

Evaluations of language models have shifted from testing syntax (e.g., model perplexity) to assessing semantics (e.g., answering multiple-choice questions)~\cite{hu2024can}. The garbling trick enables us to interpolate between these extremes and examine both aspects simultaneously. As the garbling rate increases, the LLM faces two additional tasks not present in the original evaluation. First, the LLM must infer as much as possible about the underlying text from its garbled version. Second, given that it can only partially or uncertainly recover the text, the LLM must answer the question with incomplete information. Selecting the correct answer thus requires the LLM to engage in multiple forms of reasoning\footnote{The term ``reasoning'' is used in this paper in two senses: to refer to the process by which any LLM selects an answer (as here), and to the subclass of LLMs that use test-time compute to improve inference. The meaning should be clear from context.}
concurrently.

We offer some remarks on interpreting the score curve, $s(p)$. The leftmost point, $s(0)$, corresponds to the score from the original (ungarbled) evaluation. The rightmost point, $s(1)$, can be considered the score without any meaningful context.\footnote{Note that removing the context entirely may yield a different score than providing a completely garbled context.} If a higher score indicates better performance, we would expect $s(p)$ to decrease monotonically with increasing $p$.

\section{The Contextual Core}\label{sec:core}
If the base evaluation is a multiple-choice test with $k$ options for each problem, one might assume that if the context were highly garbled, the LLM would be forced to choose a solution uniformly at random, hence
\[
\lim_{p \rightarrow 1} s(p) = \frac{1}{k}.
\]
This intuition is typically misguided. If the information in the context is publicly available—for example, if the LLM is asked a historical or scientific question—it may not require the provided context to answer. Moreover, it may be able to eliminate incorrect answers. For instance, if the answer choices are ``3,'' ``5,'' and ``10/2,'' then regardless of the context \emph{or} the question, the correct answer is likely ``3'' because we know there is a unique correct answer. Due to these effects, the score can significantly exceed $1/k$ even with a completely garbled context.\footnote{In our initial experiments with $k=3$, the best-performing LLM achieved $s(1) \approx 0.65 \gg 1/3$.}

Because one of our evaluation desiderata is to prevent performance saturation, these elevated scores can be problematic. We can obtain a more relevant (and lower) score by restricting the evaluation to problems that genuinely require the provided context. We achieve this by running the evaluation without the context and observing which problems are answered incorrectly. We refer to this subset as the ``contextual core'' and focus the evaluation on these problems. Note that this process reduces the total number of problems but increases their difficulty. If the LLM is sampling ``randomly'' (i.e., if the sampling temperature is greater than zero), one might suspect that a second run would produce an independent set of answers, leading to $s(1) = 1/k$ on the contextual core.

This intuition is \emph{also} flawed, but for subtler reasons involving that we discuss in a companion paper~\cite{bradley2024madness}. Nevertheless, restricting to the contextual core provides significantly more utility and variability in the score, in the sense that for some values of $p$, the score $s(p)$ can vary widely between models, preventing score saturation.

When comparing different models, a significant issue arises regarding the selection of the contextual core. If we use an LLM to select these problems, which LLM should we choose? Ideally, we would use an independent, high-performing LLM. However, in practice, finding such a model may be impractical, or it may be impossible to ensure that its behavior is independent of the models being evaluated. Moreover, because we are specifically selecting problems where that LLM produces incorrect answers, its performance at large values of $p$ may be lower than expected. We examine an instance of this issue in Section~\ref{sec:neosquad} and caution researchers to be mindful of this effect.

Rather than using a common set of problems in the contextual core, we could evaluate each LLM on a separate set of questions. For example, suppose we restrict each LLM to the questions it answered incorrectly at $p=1$ but correctly at $p=0$. If we measure accuracy, the score curve would drop from 1 to 0 as $p$ increases from 0 to 1, and the shape of the curve would reveal how adept the model's reasoning abilities are as the garbling worsens. While helpful for understanding a single model, this evaluation does not allow us to compare performance \emph{between} models—a typical goal in these assessments—so we do not pursue it here.

\section{NeoSQuAD Evaluation} \label{sec:neosquad}
To illustrate our method, we create a new evaluation dataset and framework, apply the garbling trick, and analyze the results across nine different LLMs. We call the new dataset ``NeoSQuAD''.

We generated a set of 10,000 multiple-choice questions by sampling from the 100,000 questions in the answerable subset of the SQuAD 2.0 dataset~\cite{rajpurkar2018know}. Each SQuAD problem consists of a paragraph or two of context followed by a question. Two-thirds of the questions are answerable, and a correct answer is a substring of the context. Since SQuAD is not a multiple-choice test, we prompted \texttt{gpt-4o-2024-08-06} to provide two additional incorrect answers, resulting in a 10,000-problem multiple-choice test with three choices per problem. We also ensured that no answer was substring of another answer. The SQuAD data is ASCII-encoded; we garbled each byte with probability $p$ by resampling a garbled byte uniformly at random from the 256 possibilities.

Next, we needed to identify the ``contextual core'' of problems for which the context is necessary. In our initial attempts, we directly prompted an LLM (``Does this question contain enough information to select an answer from the following list of options?''). This approach proved insufficient: even though the LLM claimed it didn't have enough information but was still able to answer more accurately than at random. To construct a more conservative contextual core, we instead asked the question and demanded an answer. We prompted two LLMs (\texttt{gpt-4o-2024-08-06} and \texttt{gemini-1.5-pro-001}) and restricted the evaluation to the subset of questions where both LLMs chose the wrong answer. This process produced a contextual core of 1,027 problems (approximately 10\% of our initial set).

We then generated score curves for a set of nine LLMs at garbling rates of
\[
p = [0, 0.2, 0.3, 0.4, 0.5, 0.6, 0.8, 0.9].
\]

\section{NeoSQuAD Results}

We applied the NeoSQuAD evaluation to 12 LLMs. The models were chosen for diversity— they include some of the largest and some of the smallest models. The models can be divided into 8 base LLMs and 4 ``reasoning'' LLMs (which leverage test-time compute). The traditional LLMs are \texttt{Qwen2-72B-Instruct}, \texttt{Meta-Llama-3.1-70B-Instruct}, \texttt{Meta-Llama-3.1-8B-Instruct}, \texttt{Phi-3-medium-4k-instruct}, \texttt{Mixtral-8x7B-Instruct-v0.1}, \texttt{gemini-1.5-pro-001}, \texttt{gemini-1.5-pro-002}, \texttt{gpt-4o-2024-08-06}. The reasoning LLMs are \texttt{gemini-2.5-pro-preview-03-25}, \texttt{o1-preview-2024-09-12}, \texttt{o3-mini-2025-01-31}, and \texttt{Deepseek-R1}. Note that the default reasoning effort was used for the reasoning models in all cases (e.g., \texttt{o3-mini-2025-01-31} corresponds to \texttt{o3-mini} (medium)).

\begin{figure}[htbp]
  \centering
  \includegraphics[width=1.0\textwidth]{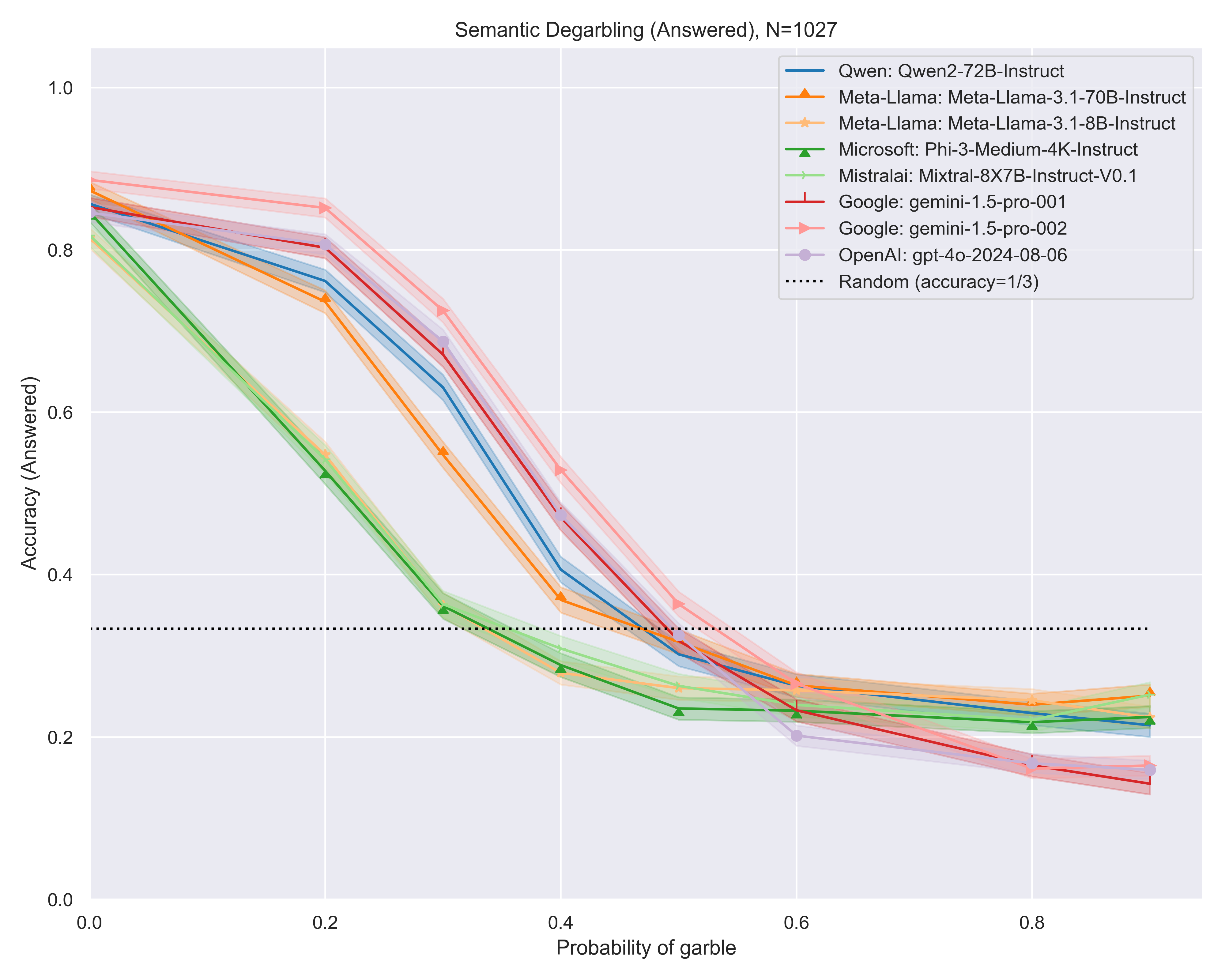}
  \caption{NeoSQuAD score curves across eight traditional (non-reasoning) LLMs, normalized by the number of questions answered instead of the number of answers parsed. The shaded region around each curve represents $\pm 1\sigma$ confidence intervals.}
  \label{fig:non_reasoning}
\end{figure}

\begin{figure}[htbp]
  \centering
  \includegraphics[width=1.0\textwidth]{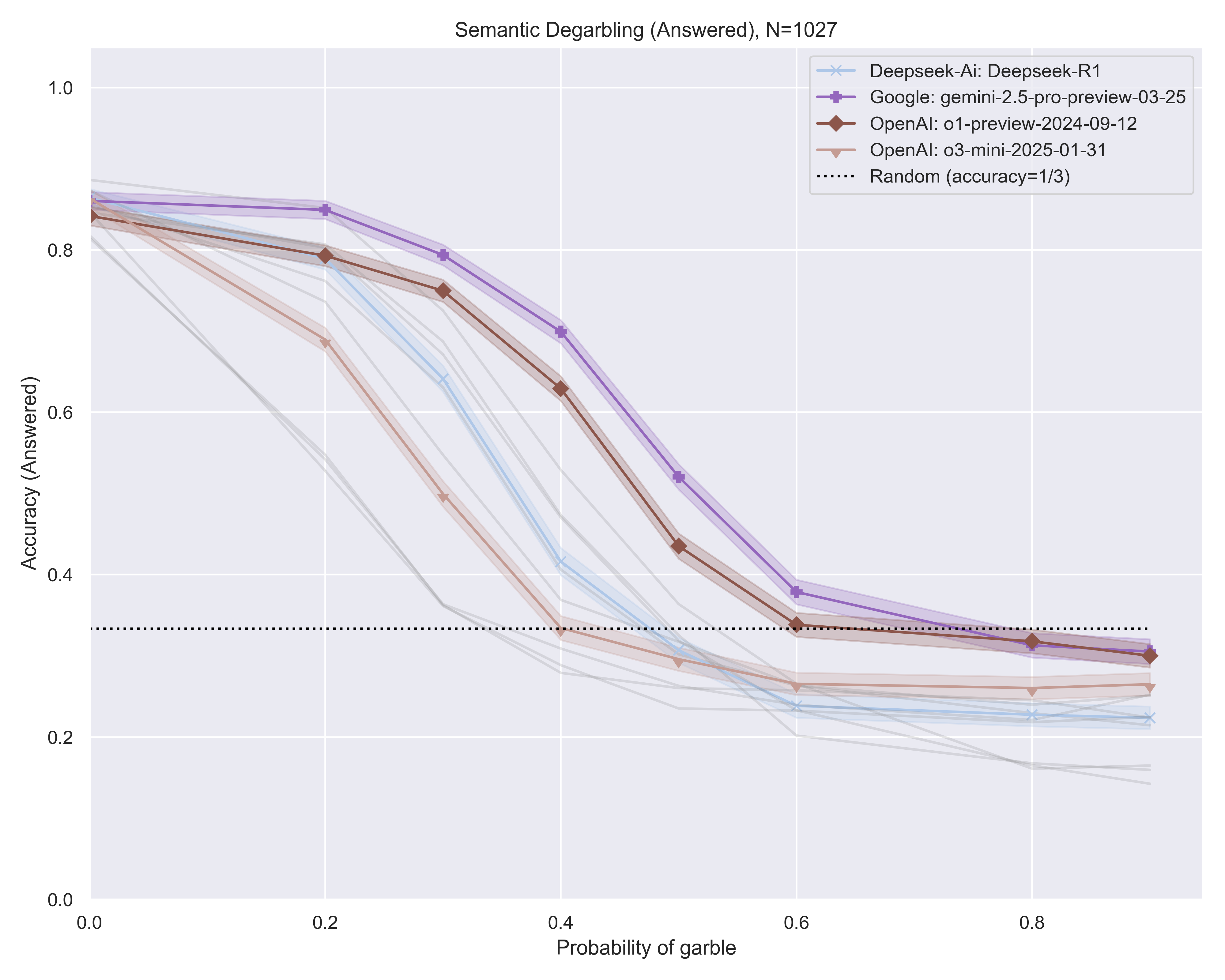}
  \caption{NeoSQuAD score curves across four reasoning LLMs, normalized by the number of questions answered instead of the number of answers parsed. The shaded region around each curve represents $\pm 1\sigma$ confidence intervals. The performance of the eight non-reasoning models is faintly shown in grey for comparison.}
  \label{fig:reasoning}
\end{figure}

The results for traditional LLMs are shown in Figure~\ref{fig:non_reasoning}.\footnote{Note that evaluator needs to handle unparseable responses; this detail is discussed in Appendix~\ref{sec:valid}.} and the results for reasoning models are shown in Figure~\ref{fig:reasoning}.

Figures~\ref{fig:non_reasoning} and~\ref{fig:reasoning} allow us to observe several general features of these models. At $p=0$ (i.e., in the original, ungarbled evaluation), model performance is hard to distinguish. The accuracy is nearing saturation, and many models fall within the confidence intervals of others, making reliable differentiation difficult. However, by $p=0.3$, model performance has largely separated, reducing saturation and providing more variable and hence more informative scores.

Qualitative behaviors also become apparent. Consider the three smaller models: Microsoft's \texttt{Phi-3-medium-4k-instruct} (14B parameters), Mistral's \texttt{Mixtral-8x7B-Instruct-v0.1} ($8 \times 7$B = 56B parameters), and Meta's \texttt{Llama-3.1-8B-Instruct} (8B parameters). These models are not particularly distinct at $p=0$ (notably, \texttt{Phi-3-medium-4k-instruct} appears identical to larger models), but when considering the full score curves, they behave similarly to each other and differently from the larger models. Interestingly, \texttt{Mixtral-8x7B-Instruct-v0.1} behaves like a smaller model, whereas \texttt{Llama-3.1-70B-Instruct} behaves like a larger model, despite their similar parameter counts (56B vs. 70B). Future research might explore the differences between dense and mixture-of-expert architectures on score curves.

The score at $p=0.4$ provides a useful ordering of the models. In decreasing order of performance:
\begin{itemize}
  \item \texttt{gemini-2.5-pro-preview-03-25}
  \item \texttt{o1-preview}
  \item \texttt{gemini-1.5-pro-001}
  \item Tied: \texttt{gpt-4o-2024-08-06}, \texttt{gemini-1.5-pro-001}
  \item Tied: \texttt{Deepseek-R1}, \texttt{Qwen2-72B-Instruct}
  \item \texttt{Llama-3.1-70B-Instruct}
  \item \texttt{o3-mini-2025-01-31}
  \item Tied: \texttt{Phi-3-medium-4k-instruct}, \texttt{Mixtral-8x7B-Instruct-v0.1}, \texttt{Llama-3.1-8B-Instruct}
\end{itemize}

Our method of constructing the contextual core also introduces some artifacts. Because we selected problems where \texttt{gemini-1.5-pro-001} and \texttt{gpt-4o-2024-08-06} are incorrect, their score curves suffer at large values of $p$. Note that \texttt{gemini-1.5-pro-002} is similar enough to \texttt{gemini-1.5-pro-001} that it experiences the same effect. This phenomenon becomes visible at $p=0.6$, leading to a distinct performance cluster where these three score curves are noticeably lower than those of other models. This contrasts with their behavior at $p \leq 0.5$, where these models outperform others.

As discussed in Section~\ref{sec:core}, since there are three multiple-choice options, one might expect model performance to converge to $1/3$ at $p=1$. However, by $p=0.6$, the accuracy has dropped significantly below $1/3$ for all models except \texttt{o1-preview} and \texttt{gemini-2.5-pro-preview-03-25}.

If we were to draw some rules of thumb from this analysis, it appears that the most informative region of the score curve lies in $p \in [0.2, 0.5]$, and particularly $p \in [0.3, 0.4]$. In this range, the garbling is sufficient to require reasoning, reduce saturation, and differentiate models, yet low enough to still observe meaningful performance differences.

Finally, these score curves provide insights into non-reasoning versus reasoning models. The two largest reasoning models, \texttt{o1-preview} and \texttt{gemini-2.5-pro-preview-03-25}, are significantly better than all other models from $p=0.3$ onward. Note that o1-preview uses a gpt-4o base model, so the improvement granted by test-time compute is particularly clear. Note also that these were the only models that achieved a score of approximately $1/3$ accuracy in the high garble-rate regime. In other words, they were able to overcome the implicit bias towards incorrect answers that all the other models suffered from.

These results also reveal two distinct clusters among the reasoning models under increasing garbling.  In particular, OpenAI's \texttt{o1-preview} and \texttt{o3-mini} exhibit markedly different behavior: \texttt{o1-preview} remains robust even at high corruption rates, while \texttt{o3-mini} degrades nearly in step with non-reasoning baselines.  Likewise, Google's \texttt{Gemini-2.5-Pro} joins \texttt{o1-preview} in the top tier, whereas DeepSeek's R1 tracks with the weaker group.

One possible explanation is model capacity, i.e., the number of active parameters engaged during inference.  Because OpenAI and Google treat model parameters as proprietary, the true number of active parameters is not publicly known, so the following suggestions should be treated as highly speculative. Public reports suggest that \texttt{o1-preview} employs around 300 billion active parameters from a presumably much larger model~\cite{ben_abacha2024medec}.  In contrast, \texttt{o3-mini} is a 200 billion-parameter dense model~\cite{sakthi2025o3mini-deepseek}. Despite a total of 671 billion parameters in its Mixture-of-Experts architecture, DeepSeek's R1 activates only about 37 billion at a time.  Earlier Gemini releases reportedly used 1.56 trillion total parameters, so \texttt{Gemini-2.5-Pro} likely maintains a similarly large active footprint~\cite{saxena2023colossus}.  Perhaps, then, the poorer performance of \texttt{o3-mini} and R1 may be related to their seemingly more limited active parameter count.

\section{Conclusions and Extensions}

We have introduced the ``garbling trick,'' a novel technique for evaluating LLMs that transforms an existing evaluation into a series of more challenging tests. These new tests can mitigate score saturation and provide richer insights into the reasoning abilities of LLMs.

Several potential extensions are worth mentioning. First, while this paper focused on multiple-choice evaluations with a ``context + question + answers'' structure, the garbling trick can be applied more broadly. For example, many multiple-choice problems do not clearly separate the context from the question. In such cases, we might garble the entire context and question but leave the answers ungarbled. Second, most LLMs have a controllable ``temperature'' parameter: setting it to zero produces the maximum likelihood response, but setting it to a positive value allows for resampling of responses. Investigating the effect of the temperature parameter on $s(p)$ would be interesting; it may reduce performance at small values of $p$ but increase it at larger values.

\bibliographystyle{plain}
\bibliography{references}

\appendix
\section{Valid Answers}\label{sec:valid}
\begin{figure}[htbp]
  \centering
  \includegraphics[width=1.0\textwidth]{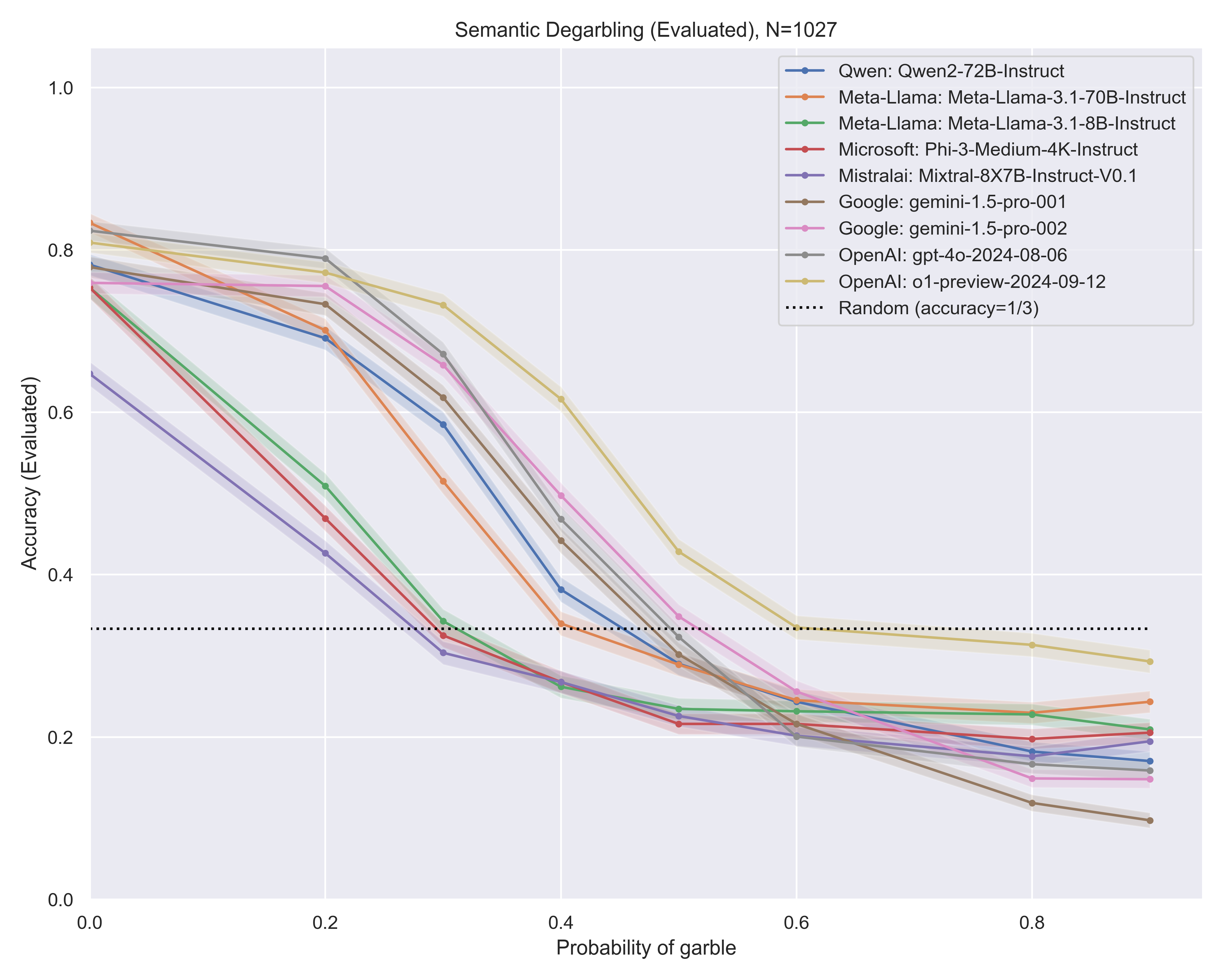}
  \caption{NeoSQuAD score curves normalized by the number of questions asked instead of the number of answers parsed.}
  \label{fig:score_curve_asked}
\end{figure}

\begin{figure}[htbp]
  \centering
  \includegraphics[width=1.0\textwidth]{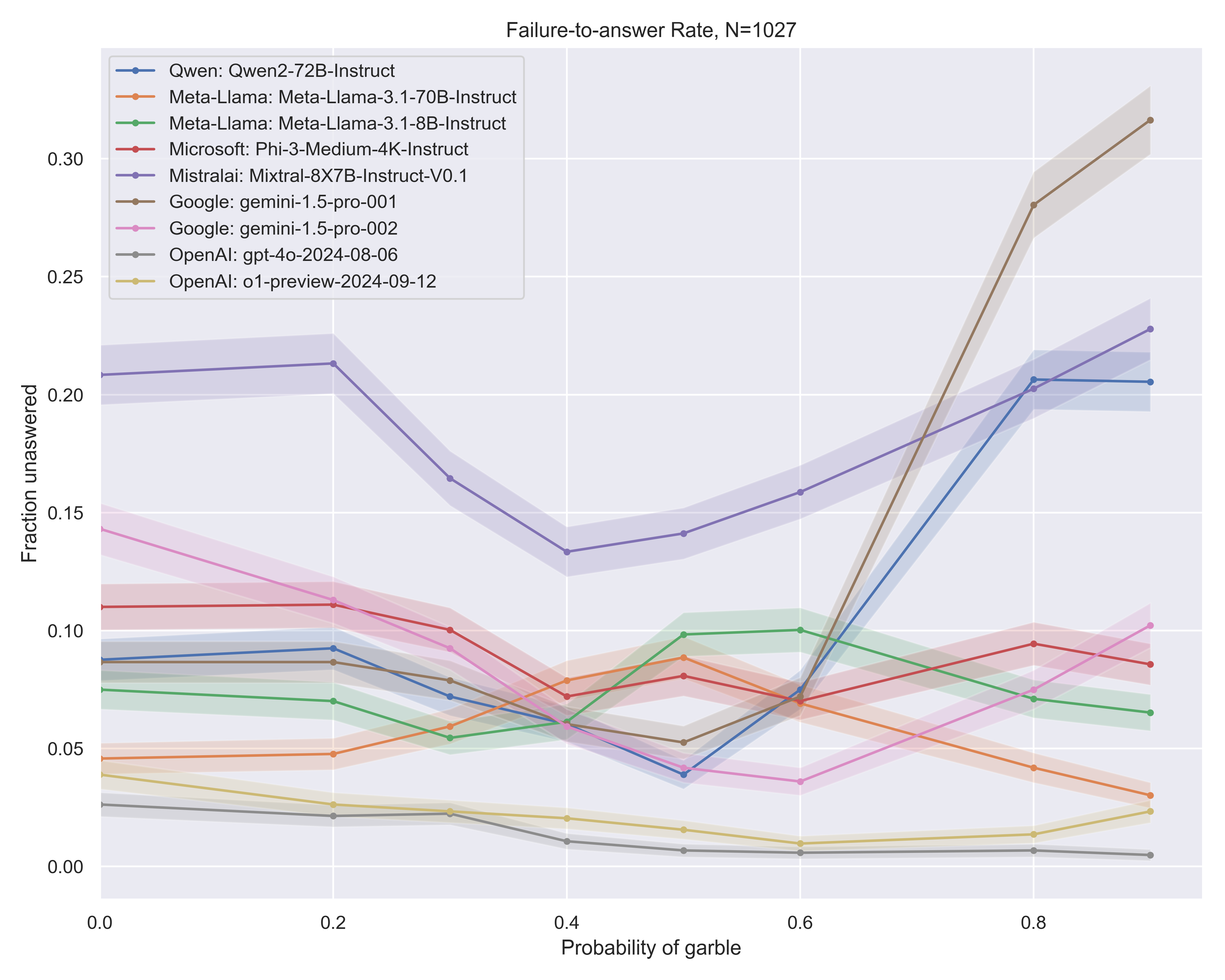}
  \caption{Rate of invalid answers while computing NeoSQuAD score curves.}
  \label{fig:invalid_rate}
\end{figure}

An LLM may fail to provide a valid answer to a question. An invalid answer may occur because the framework expects a certain structure (such as a parseable JSON blob) that the response violates, or because the question triggers a safety mechanism and the LLM refuses to answer. In the NeoSQuAD results reported in Section~\ref{sec:neosquad}, we considered ``accuracy'' as the ``number of correct answers'' divided by the ``number of valid answers.'' Alternatively, we can define accuracy using the stricter criterion of ``number of correct answers'' divided by the ``number of questions asked.'' We present the resulting score curves in Figure~\ref{fig:score_curve_asked} and plot the fraction of invalid answers per LLM in Figure~\ref{fig:invalid_rate}.

The ``safety violation'' issue will likely persist, but most invalid answers resulted from JSON parse failures. OpenAI has recently added a ``Structured Output'' option that enforces valid output, and presumably other providers will follow suit, which should largely eliminate parse failures. Since ``safety violations'' constitute a small fraction of invalid answers, we anticipate that invalid responses will not be a significant concern in the long term.

\end{document}